\documentclass{article}
\usepackage[english]{babel}

\usepackage[letterpaper,top=2cm,bottom=2cm,left=3cm,right=3cm,marginparwidth=1.75cm]{geometry}

\usepackage{amsmath}
\usepackage{amssymb}
\usepackage{graphicx}
\usepackage[utf8]{inputenc}
\usepackage[T1]{fontenc}
\usepackage{fancyvrb}
\usepackage{listings}
\usepackage{subcaption}
\usepackage{booktabs}
\usepackage{authblk}

\usepackage[colorlinks=true, allcolors=blue]{hyperref}

\title{Coreset selection based on Intra-class diversity}
\author[1]{Imran Ashraf}
\author[2]{Mukhtar Ullah}
\author[3]{Muhammad Faisal Nadeem}
\author[4]{Muhammad Nouman Noor}
\affil[1]{NUCES-FAST, Department of Computer Science, Islamabad, 44000, Pakistan}
\affil[2]{NUCES-FAST, Department of Electrical Engineering, Islamabad, 44000, Pakistan}
\affil[3]{Informatics Complex, Islamabad, 46000, Pakistan}
\affil[4]{NUCES-FAST, Department of Artificial Intelligence and Data Science, Islamabad, 44000, Pakistan}

\begin{document}
\maketitle

\begin{abstract}
In recent years, Deep Learning (DL) models have transformed countless domains, including the healthcare sector. In particular, these models have generated impressive outcomes in biomedical image classification by learning intricate features and enabling accurate diagnostics pertaining to complex diseases. More recently, studies have adopted two different approaches to train DL models: training from scratch and transfer learning. In the latter case, pre-trained DL models are used as a backbone, while a few top layers are fine-tuned to customize them for a specific task. Nonetheless, both these approaches demand substantial computational time and resources due to the involvement of massive datasets in model training. These computational demands are further increased due to the design-space exploration required for selecting optimal hyperparameters, which typically necessitates several training rounds. With the growing sizes of biomedical datasets, exploring solutions to this problem has recently gained the research community's attention. A plausible solution is to select a subset of the dataset for training and hyperparameter search. This subset—referred to as the corset—must be a representative set of the original dataset. A straightforward approach to selecting the coreset could be employing random sampling, albeit at the cost of compromising the representativeness of the original dataset. A more critical limitation of random sampling is the bias towards the dominant classes in an imbalanced dataset. Even if the dataset has interclass balance, this random sampling will not capture intraclass diversity. The proposed study addresses this issue by introducing an intelligent, lightweight mechanism for coreset selection. Specifically, it proposes a method to extract intraclass diversity, forming per-class clusters that are then used for the final sampling. We demonstrate the efficacy of the proposed methodology by conducting extensive classification experiments on a well-known biomedical imaging dataset, the Peripheral Blood Cell (PBC) dataset. These experiments demonstrate that the proposed method of intelligently selecting a coreset of the complete dataset outperforms the random sampling approach on several performance metrics for uniform conditions. Finally, we recommend that the proposed study transform learning-based R\&D domains by saving extensive computational resources and time. 
\end{abstract}

\section{Introduction}

In recent years, Deep Learning (DL) models have exhibited great potential in the healthcare sector 
~\cite{chakraborty2024machine, chen2020deep, miotto2018deep}. These models are trained on large datasets of biomedical images, such as CT scans, MRIs, and X-rays, to accurately detect abnormalities and complex patterns that can lead to a diagnosis or prediction of diseases ~\cite{haque2020deep, chakraborty2020overview, kim2020deep}. Consequently, the pervasive use of DL models in this area has transformed the detection of diseases, exploration of patient-centric treatment options, and improvement of diagnostics ~\cite{suzuki2017overview}. Similarly, DL models have demonstrated exceptional performance in personalized treatment by analyzing genomic data and offering customized drug infusions and interventions ~\cite{roy2025ai}. At the core of DL algorithms in the healthcare domain is the well-known medical image classification technique, a powerful tool in many computer-aided diagnoses. In this case, a medical image is input into a DL model, such as a Convolutional Neural Network (CNN), which extracts various features from the image and returns a corresponding class label ~\cite{sarvamangala2022convolutional}. When a CNN is trained on a massive dataset of labeled images, it offers remarkable accuracy. Consequently, improved treatment planning, quicker diagnosis, and better patient outcomes are possible.

Primarily, there are two methods for training a DL model for medical image classification. In the first method, these models are trained from scratch, typically requiring large datasets and computational resources. Furthermore, they necessitate the meticulous selection of CNN architectures and hyperparameter tuning, which in turn demands additional resources and time. Despite their advantage of complete control and customized performance, they are limited by the requirement of large labeled datasets and computational resources. Therefore, a compelling alternative is transfer learning models, where a DL model is pre-trained on a specific task and can be tailored to perform several other related tasks ~\cite{gupta2022deep}. In transfer learning, the top few layers of a CNN architecture can be modified to customize the model for a specific task. Due to its pre-training, this DL model requires less computational time and resources, as well as a shorter convergence time, than a model developed from scratch ~\cite{salehi2023study, hussain2019study}. However, these models are limited due to a lack of flexibility and issues with hyperparameter tuning.

With the growing sizes of biomedical datasets, exploring innovative solutions to DL model training has recently garnered the research community's attention ~\cite{salehi2023study}. A plausible solution is to select a subset of the dataset for training and tuning the hyperparameters. This subset—also known as the \textit{corset}—must be a representative set of the original larger dataset ~\cite{chai2023efficient, hong2024evolution, yang2023towards}. If this coreset can help train a CNN model with comparable accuracy and other performance metrics, a significant amount of computational resources and time can be saved. 

A trivial yet straightforward approach to selecting the coreset could be using random sampling. However, a random selection of the original dataset can compromise its representativeness and complete coverage. Another issue is the bias towards the dominant classes in the case of an imbalanced dataset. Even if the dataset comprises interclass balance, the random sampling cannot capture intraclass diversity, leading to poor generalization, overfitting, and low robustness ~\cite{guo2022deepcore}.

Recently, several studies have focused on coreset selection ~\cite{chai2023efficient, hong2024evolution, yang2023towards}. Some of these works employ uncertainty-based coreset selection, using Bayesian models and entropy techniques to select images from a large dataset ~\cite{campbell2018bayesian}. Similarly, some studies focus on gradient-based coreset selection, feature representation, and domain-specific techniques ~\cite{bachem2017practical, huang2021novel, hong2024evolution}. Other diversity-driven studies utilize clustering, k-center greedy algorithms, and the Determinantal Point Process (DPP) to select representative coresets ~\cite{chai2023efficient, ding2019greedy, tremblay2019determinantal}. All these methods represent preliminary work in coreset selection and are limited by computational costs, a lack of generalization, and limited intraclass diversity. 

Therefore, this study introduces an innovative, lightweight mechanism for coreset selection by extracting intraclass diversity to form per-class clusters, thereby completing the final sampling process. We demonstrate the effectiveness of this method by performing several classification experiments on a popular biomedical imaging dataset, Peripheral Blood Cell (PBC) ~\cite{pbc-dataset}. This dataset is a labeled set of images of microscopic blood cells, used for training and evaluating models that categorize blood cells. The PBC dataset contains eight different categories of blood cells. The goal is to develop a model to identify a specific class of blood cells in the PBC dataset.

In this work, we demonstrate the efficacy of the proposed intraclass diversity technique in classifying blood cells using the PBC dataset. Specifically, we create the following two models:

\begin{itemize}
    \item A DL model from scratch and train it on the original PBC dataset.
    \item Use a pretrained ResNet model as a backbone and employ the transfer learning technique to perform image classification using the PBC dataset.
\end{itemize}

Next, we choose two subsets from the original PBC dataset. First, the Random Sampling (RS) method randomly selects the coreset from the original PBC dataset. In the second scenario, we use the Intelligent Sampling (IS) method to finalize a coreset based on the proposed scheme. Subsequently, the two DL models are trained and validated on these two coresets. We show that the proposed IS methodology, which intelligently selects a coreset from the original dataset, outperforms the RS approach on multiple performance metrics under uniform conditions. Furthermore, the proposed method results in reduced computational complexity.

To the best of our knowledge, none of the state-of-the-art studies have utilized intraclass diversity for coreset selection. We believe that the proposed work can help transform DL-based research and industrial applications by saving extensive computational resources. 

The remainder of this paper is organized as follows: Section 2 describes a comprehensive literature review of the related works in this field, concluding that the proposed method is innovative and has not been discussed in any prior study. Section 3 explains the proposed IS method using intraclass clustering. Section 4 presents the experimental setup and results using the two DL models mentioned above. Finally, Section 5 provides the conclusions. 

%%%%%%%%%%%%%%%%%%%%%%
\section{Related Work}

With the emergence of massive datasets in all real-world fields, it is essential to explore innovative solutions for training state-of-the-art, efficient models. In many cases, extracting a representative subset, or coreset, from a large dataset can significantly reduce computational costs and the time required to train advanced Deep Learning (DL) models. However, giant datasets might not exist in some industries, requiring a transition from big data to good data. Prof. Andrew Ng notably proclaimed, "\textit{In many industries where giant data sets simply do not exist, I think the focus has to shift from big data to good data. Having 50 thoughtfully engineered examples can be sufficient to explain to the neural network what you want it to learn}." In both scenarios, it is essential to seek innovative methods for selecting a coreset. In data-intensive applications such as biomedical image analysis, training a model on large volumes of data is computationally expensive. Therefore, selecting a coreset to extract a representative subset of the original dataset can become a vital step in minimizing computational costs.

Several studies have addressed the coreset selection problem in recent years. Traditionally, these subsets were chosen by representing learned features. For instance, a common technique has been to use Principal Component Analysis (PCA) to select a subset of data that describes the most considerable variance ~\cite{sener2017active}. Although PCA has been an efficient method for coreset selection, it could still overlook various crucial patterns in image datasets. Other studies employed Self-Supervised Learning (SSL) mechanisms to produce representative subsets from unlabelled datasets. For example, they create embeddings to explore clusters and select diverse samples. In \cite{chen2020simple}, the authors propose a simple framework, SimCLR, for contrastive learning on image datasets. Their work minimizes the dependency on labelling and provides excellent outcomes in various biomedical imaging datasets. Another similar study illustrates the concept of Momentum Contrast (MoCo) for unsupervised image subset selection by developing an on-the-fly dictionary that supported unsupervised learning ~\cite{he2020momentum}. These works have been criticized for overlooking intra-class diversity in their coreset selection process, which is critical in biomedical imaging.

Another important coreset selection methods are gradient-based techniques, which retain those samples whose gradients can estimate the gradient of the entire dataset during model training. In ~\cite{koh2017understanding}, the researchers leveraged influence functions to approximate the impact of each sample on model parameters to specify a representative subset. Likewise, another study provides a remarkable example of using a bilevel optimization technique to select samples that can generalize optimally on a validation dataset ~\cite{killamsetty2021glister}. While these gradient-based methods have demonstrated excellent performance for large-scale datasets, their high computational cost limits their practical application in biomedical image analysis. Another concern is their neglect of intraclass diversity, which leads to bias in coreset selection. 

As diagnostics and clinical studies cannot ignore unique cases due to the considerable risks involved, recent works have addressed the issue of intraclass diversity in coreset selection. For instance, authors in \cite{guo2022margin} address the issue of unique data points in biomedical image datasets and propose a margin-aware intraclass novelty identification method, which identifies any distinct samples within a class in an imaging dataset. They conclude that a DL model trained with their coreset will be able to detect any rare diseases using an X-ray image dataset. Although they use a novelty-based approach, their coreset selection is limited due to domain-specific assessments of margins in the dataset. Furthermore, they struggle with issues like scalability and generalizability.

Some recent studies have tackled the generalization issue by introducing inter-observer variability in biomedical image datasets ~\cite{quinn2023interobserver, schmidt2023probabilistic, benchoufi2020interobserver}. These studies identify cases where expert labeling disagrees, enabling them to highlight clinically essential cases. Although these works are significant, they rely on annotated datasets that might not always be available. Moreover, despite their usefulness, they may still be unable to generalize their outcomes across diverse datasets.

Traditional methods use interclass diversity in coreset selection, which can significantly influence the accuracy of a model. These methods often overlook the distinctive features within intraclass data points by disregarding intraclass diversity. In ~\cite{zhou2016feature}, the researchers offered a solution by introducing the Interclass and Intraclass Relative Contributions of Terms (IIRCT) method, which incorporates both inter- and intra-class data points in the feature selection process. The IIRCT method was evaluated on text classification tasks, and its viability in medical imaging datasets was not explored. Nonetheless, their approach showed promising results by evaluating intraclass variability. Similarly, a study by Kaushal \textit{et al.} utilized intraclass coreset selection in dynamic learning scenarios in computer vision ~\cite{kaushal2018learning}. They reported improved performance results with fewer training data on complex computer vision tasks.

In biomedical imaging, very few studies have worked on intraclass diversity. For instance, Georgescu et al. described a method for promoting diversity in medical image categorization ~\cite{Georgescu2023}. In 2024, a recent work introduced an innovative medical imaging strategy called Evolution-aware VAriance (EVA) coreset. This study preserves the intraclass diversity in data points and improves categorization efficiency ~\cite{hong2024evolution}.

Various datasets, such as the Peripheral Blood Cell (PBC) dataset and other imaging datasets, like X-ray image collections, are notable repositories for performing coreset selection experiments ~\cite{pbc-dataset}. A practical study was conducted by Seneer and Savarese, in which they implemented coreset selection for training a CNN model, significantly reducing annotation needs without compromising accuracy ~\cite{sener2017active}. Another clustering-based study developed a coreset selection framework that preserves diversity ~\cite{bachem2017practical}. These methods have demonstrated their efficacy in general-purpose applications; however, they require customization to observe unique patterns for disease prediction and diagnostics. 

Despite numerous recent works on coreset selection, most of these studies fail to address intraclass diversity, resulting in limited accuracy. Some proposed methods, such as gradient-based techniques, require large-scale computational resources. Consequently, they pose scalability issues in complex biomedical imaging datasets. Others rely on embeddings from their pre-trained models, which can introduce inherent biases. In general, intraclass diversity has not been appropriately addressed in any study. 

A significant issue in this field is a lack of adequate evaluation methods for assessing the efficacy of the coreset. As a result, it becomes challenging to perform reasonable comparisons. For these reasons, the proposed study offers a unique solution to introduce an Intelligent Sampling (RS)-based intraclass diversity method in the coreset selection process. For comparison, we use Random Sampling (RS) as a baseline method for evaluating our work. This technique yields promising results in imbalanced datasets with skewed class distributions. Due to its adequate representativeness, it can be employed in diverse institutions and datasets, addressing the generalizability issues in traditional methods.

%%%%%%%%%%%%%%%%%%%%%%
\section{Problem Definition}

Given a large labeled training set $\mathcal{T} = \{(x_i, y_i)\}_{i=1}^{N}$ of size $N$. In the context of supervised learning, $x_i$ are viewed as features from an input space $\mathcal{X}$ whereas $y_i$ as the class labels from an output space $\mathcal{Y}$. 
In statistical machine learning a training set like $T$ is assumed to come from an unknown probability distribution.  The aim here is to build a classifier \( f_\theta: \mathcal{X} \to \mathcal{Y} \), parameterized by \( \theta \), which can be trained by minimizing a loss function \( \ell(\cdot, \cdot) \), such as cross-entropy. Here $f_\theta$ designates a specific model from the class $f$ of models. The \textbf{coreset selection problem} aims to find a subset $\mathcal{S} \subset \mathcal{T}$ with $|\mathcal{S}| \ll |\mathcal{T}|$, such that the classifier \( f_{\theta_{\mathcal{S}}} \) trained on \( \mathcal{S} \), achieves generalization performance closer to the classifier\( f_{\theta_{\mathcal{T}}} \) trained on the full dataset \( \mathcal{T} \).

%%%%%%%%%%%%%%%%%%%%%%
\section{Proposed Methodology}

\begin{figure}
\centering
\includegraphics[width=0.85\textwidth]{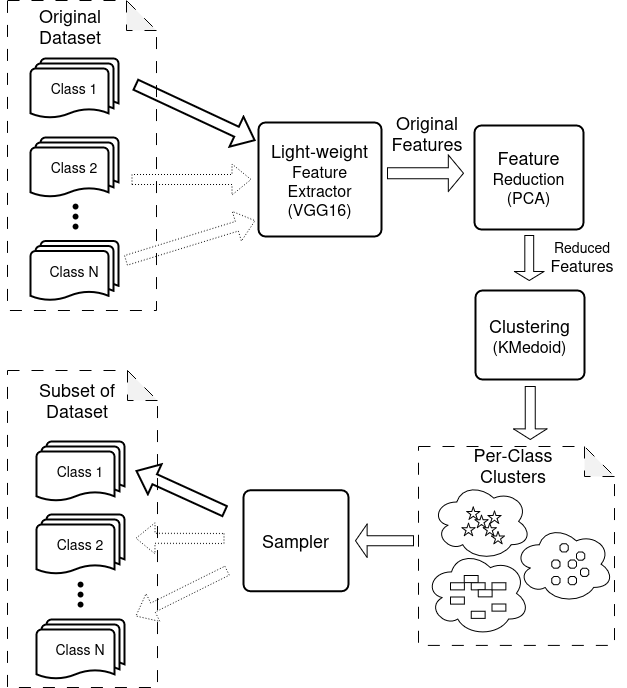}
\caption{Proposed Methodology}
\label{fig:methodology}
\end{figure}

The proposed technique leverages the intraclass diversity within the PBC dataset to create clusters of similar samples. Subsequently, it intelligently samples these clusters to represent every diverse data point within the sample. Figure~\ref{fig:methodology} depicts the block diagram of the proposed method. Primarily,  this method enables feature extraction through the VGG16 extractor. Initially, it extracts full features through the extractor. Subsequently, we use the PCA method to perform feature reduction. These reduced features are then input into the $K$-Medoids clustering algorithm, which provides per-class clusters. Finally, we create a subset of the original dataset using a sampler. Each of these steps is elaborated on in detail in the following sections.

\subsection{Feature Extraction}
In the first step, features are extracted from the samples of a single class within the PBC dataset. Since feature extraction is a computationally expensive process, we utilized a lightweight network, VGG16, a deep CNN with 16 layers, which is pretrained on the ImageNet dataset. Instead of using the final fully connected classification layer, features from the last pooling layer are extracted to obtain a compact representation of each image VGG16~\cite{vgg16}.

VGG16 transforms an input image $X$ of height $H$, width $W$, and channel size $C$ through a sequence of convolutional and pooling layers, resulting in a feature map $F\in \mathbb{R}^{d}$ at the final pooling stage:

\begin{equation}
F = f_{\theta}(X),
\end{equation}

where $f_{\theta}$ is the VGG16 network parameterized by $\theta$.

%$F$ is the extracted feature vector,
%d$ is the embedding dimension.

 % d = 4096 for the fc1 layer
 % d = 512 for the avg-pooling layer
 
For a dataset of $N$ images ${X_1, X_2, ..., X_N}$, we construct an embedding matrix $E\in \mathbb{R}^{N \times d}$ from $f_{\theta}(X_i)$ as columns

\begin{equation}
E = \begin{bmatrix} f_{\theta}(X_1) \ f_{\theta}(X_2) \ \hdots \ f_{\theta}(X_N) \end{bmatrix}^{\intercal}.
\end{equation}
where each row specifies the embedding of an image.

\subsection{Feature Reduction}
In this phase, we apply Principal Component Analysis (PCA) as a dimensionality reduction technique to further enhance computational efficiency and reduce redundancy in the extracted embeddings. This step is especially significant because we plan to use $K$-Medoids clustering in the next step. 

PCA is a statistical technique that transforms the high-dimensional embedding matrix $E$ into a lower-dimensional subspace while preserving the most significant variance in the data.

Given the embedding matrix $E \in \mathbb{R}^{N \times d}$, PCA projects the embeddings onto a new basis formed by the principal components. This transformation has been formulated as follows:

\begin{equation}
E_{\text{red}} = E W_k,
\end{equation}

where,

$E_{\text{red}} \in \mathbb{R}^{N \times k}$ denotes the reduced embedding matrix,

$W_k \in \mathbb{R}^{d \times k}$ is the matrix containing the top $k$ eigenvectors of the covariance matrix $C = \frac{1}{N} E^T E$.

The principal components are selected to maximize the variance retained in the reduced space, ensuring minimal loss of information. The explained variance ratio is formulated as follows:

\begin{equation}
\lambda_k = \frac{\sum_{i=1}^{k} \sigma_i^2}{\sum_{j=1}^{d} \sigma_j^2},
\end{equation}

where $\sigma_i^2$ specifies the eigenvalues of the covariance matrix $C$. The value of $k$ is chosen based on the cumulative explained variance threshold, typically retaining $95\%$ of the total variance.

Through this step, we obtain a more compact representation of image embeddings by applying PCA, resulting in computationally efficient clustering.

\subsection{Intraclass Clustering}
This clustering phase is the most critical step in the proposed method. In this step, we utilize $K$-Medoids clustering technique to form intraclass clusters. The complexity of $K$-Medoids is $O(N^2KT)$ where $N$ is the number of samples, $T$ is the number of iterations, and $K$ is the number of clusters. Here is the rationale behind using feature reduction in the last step.

In contrast to the k-means algorithm, $K$-Medoids chooses actual data points as centers, known as medoids or exemplars. Consequently, it allows for greater interpretability of the cluster centers than in k-Means, where the center of a cluster is not necessarily one of the input data points. Instead, it is the average between the points in the cluster. Another advantage of $K$-Medoids is that it minimizes the sum of pairwise dissimilarities instead of the sum of squared \textit{Euclidean} distances, making it more robust to noise and outliers compared to k-Means.

A vital parameter to choose for K-Medoid clustering is the number of clusters. We use the \textit{Silhouette} score to find the optimal number of clusters. In the following sequel, we write $i \in C_I$ to denote instance $i$ in cluster $C_I$. Similarly, $j \in C_J$ specifies instance $j$ in cluster $C_J$. The Silhouette value quantifies how similar an object is to its own cluster (cohesion) compared to other clusters (separation). The Silhouette value $s_{i}$ of a single instance is given as follows:

\[
s_i = \frac{b_i - a_i}{\max(a_i, b_i)}
\]

\noindent where the cohesion $a_i$ is computed by
\[
a_i = \frac{1}{|C_I| - 1} \sum_{i \neq j \in C_I} d_{i,j}
\]

\noindent and the separation $b_i$ by

\[
b_i = \min_{J \neq I} \frac{1}{|C_J|} \sum_{j \in C_J} d_{i,j}
\]

Here, $d_{i,j}$ is the distance between clusters $i$ and $j$. The Silhouette value ranges from -1 to +1, where a high value indicates that the object is well matched to its own cluster and poorly matched to neighboring clusters, and vice versa.

%-------------------------------
\subsection{Sampling}
Coreset selection can be defined as a probabilistic sampling task, where representative subsets are selected from a distribution to approximate the original data. We consider that $x$ is a sample from the probability distribution of a random variable $X$ if it satisfies the equation $F_X(x)=u$, where $F_X$ is the Cumulative Distribution Function (CDF) of $X$ and $u$ is a random sample from the standard uniform distribution \cite{Ullah2011}. Stated otherwise, if $u$ is a uniform random number selected from the unit interval $[0,1]$, then the $u$-th quantile of the distribution of $X$ is a sample of $X$.

The probability distribution pertinent to the coreset-selection problem is the so-called multinomial distribution \cite{pml1Book}. The joint probability mass function of random variables $X_1,X_2,\ldots,X_K$ taking respective values $n_1,n_2,\ldots,n_K$ is defined by
\[P\{X_1=n_1,X_2=n_2,\ldots,X_K=n_K\} = \frac{K!}{n_1!n_2!\cdots n_K!}p_1^{n_1}p_2^{n_2}\cdots p_K^{n_K}\]
where $n_k$ are nonnegative integers satisfying $\sum_{k=1}^{K}n_{k}=N_S$ and $p_k\ge 0$ are probabilities satisfying $\sum_{k=1}^{K}p_{k}=1$. In our present setting of the coreset-selection problem, we perceive the random variable $X_k$ as the number of examples from cluster $k$ among $|\mathcal{S}|$ samples randomly drawn from the training dataset of size $N$. From the classical interpretation of probability, $p_k$ will be proportional to the size of cluster $K$. This proportionality essentially undermines our expectation of uniform sampling, as the numbers $n_k$ will be determined by $p_k$. That, in turn, implies that underrepresented clusters will have a lower representation in the coreset relative to overrepresented clusters. 

\begin{figure}
\centering
\includegraphics[width=0.85\textwidth]{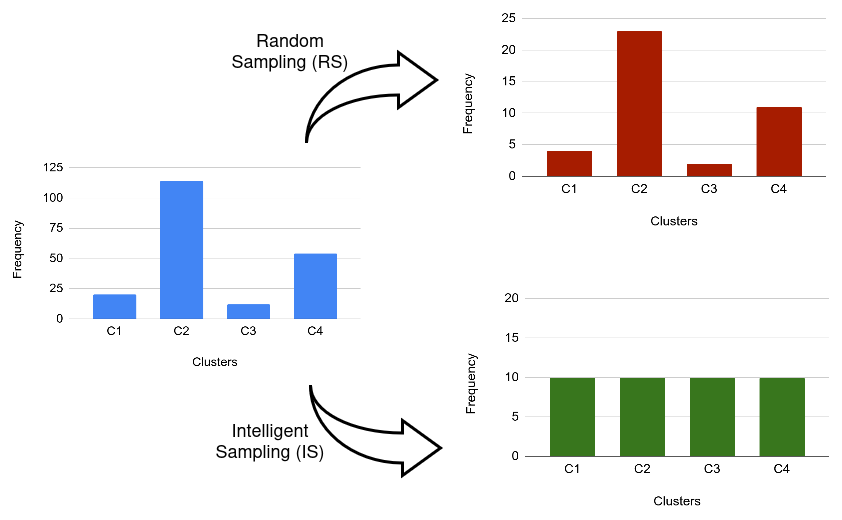}
\caption{Illustration of Random and Intelligent Sampling Methods}
\label{fig:sampling}
\end{figure}

To mitigate the above problem, our Intelligent Sampling (IS) approach treats each cluster as a separate dataset from which to sample. When sampling randomly from cluster $k$, each example is equally likely to be picked. In other words, uniformity representation is enforced by the separation of sampling problems. The efficacy of the proposed IS methodology has been illustrated in Figure~\ref{fig:sampling}. The dataset on the left contains $200$ samples. We aim to identify a coreset of this dataset, which is $5\times$ smaller (comprising only $40$ samples). Random sampling will result in a coreset, shown as the top-right distribution in this figure. In this case, the clusters that are underrepresented (\textit{e.g.,} C3) will remain underrepresented in the coreset. On the other hand, the IS method will result in a bottom-right distribution, where the total count of samples is still $40$. However, underrepresented classes have proper representation, resulting in improved learning and generalization of the model.

%%%%%%%%%%%%%%%%%%%%%%
\section{Experimental Results}
This section presents detailed experimental results obtained by applying the proposed methodology to the PBC dataset. First, we describe the dataset and its characteristics, followed by performance metrics and experimental setup. Next, we have elaborated on various results to validate the efficacy of the proposed method.

%-------------------------------
\subsection{Dataset Summary}

\begin{figure}
\centering
\includegraphics[width=0.75\textwidth]{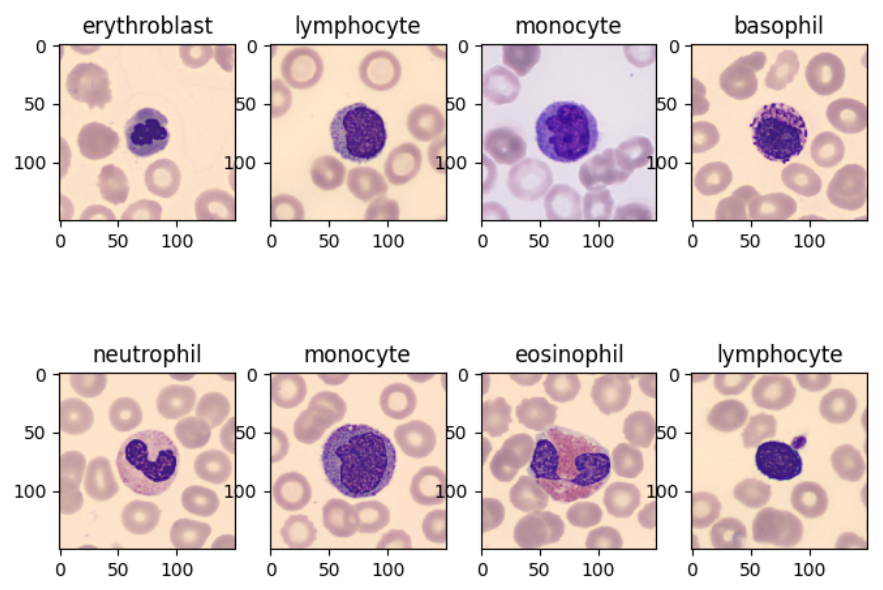}
\caption{Dataset samples}
\label{fig:dataset}
\end{figure}

\begin{figure}
\centering
\includegraphics[width=0.75\textwidth]{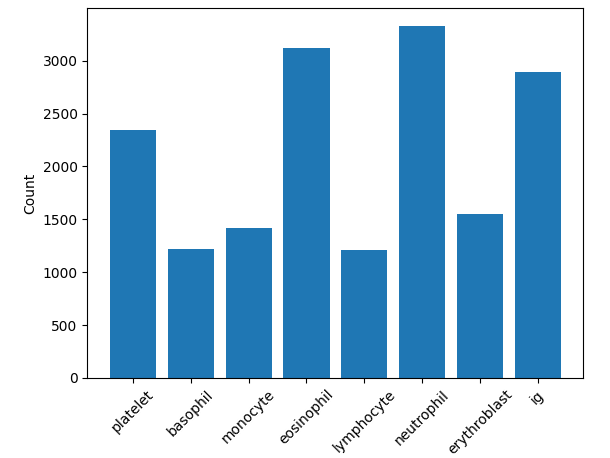}
\caption{Class Histogram}
\label{fig:histogram}
\end{figure}

To determine the efficacy of our proposed method, we conducted multiple classification experiments using the PBC dataset – a publicly available dataset containing labeled biomedical images of microscopic blood cells ~\cite{pbc-dataset}. Primarily, the PBC dataset contains eight different classes of these microscopic cells and has been used in several recent classification studies. The dataset contains 17092 labeled images acquired using the \textit{CellaVision} DM96 analyzer in the Core Laboratory at the Hospital Clinic of Barcelona. Figure~\ref{fig:dataset} depicts the samples of various PBC classes, including erythroblast, lymphocyte, monocyte, basophil, neutrophil, monocyte, eosinophil, and lymphocyte. Figure~\ref{fig:histogram} presents the distribution of the dataset in these classes, showing their counts in a typical blood sample. It can be observed that basophils have the lowest count, whereas neutrophils have the highest count. Each image in the dataset has a resolution of $360\times363$ pixels in JPG format. Expert pathologists have labeled these images after ensuring that the samples were taken from individuals without hematologic or oncologic infections. Additionally, these individuals were not undergoing any pharmacological treatment at the time of blood collection.

%-------------------------------
\subsection{Experimental Setup}
The experimental setup for conducting all simulation experiments using the conda environment is presented in the following table.

% \begin{verbatim}
\begin{Verbatim}[samepage=true, fontsize=\small]
Python implementation   : CPython
Python version          : 3.9.18
sklearn                 : 1.3.2
TensorFlow              : 2.14.0
Numpy                   : 1.26.1
Pandas                  : 2.1.2
Matplotlib              : 3.8.1
Seaborn                 : 0.13.0
\end{Verbatim}
% \end{verbatim}

We developed two DL models for simulation experiments: a model built from scratch and a pretrained ResNet model customized to perform image classification using the PBC dataset. Subsequently, we chose two coresets through random sampling and the proposed intelligent sampling methods and trained and validated both DL models using these coresets. We evaluated the performance of both models on these coresets and compared them in terms of clustering accuracy and computational time. The results are presented in the following subsections. 

We evaluated both models by splitting the dataset into a 70/30 ratio, where $70\%$ of the images from the dataset belonged to the eight classes for training and $30\%$ of the images were reserved for testing the models.

%-------------------------------
\subsection{Performance Metrics}

We mainly used four performance metrics to evaluate the effectiveness of the proposed method. These metrics include accuracy, precision, recall, and F1 Score. These are calculated using the following equations:
\begin{equation}
\text{Accuracy} = \frac{\mathrm{TP} + \mathrm{TN}}{\mathrm{TP} + \mathrm{TN} + \mathrm{FP} + \mathrm{FN} }    
\end{equation}

\begin{equation}
\text{Precision} = \frac{\mathrm{TP}}{\mathrm{TP} + \mathrm{FP}}
\end{equation}

\begin{equation}
\text{Recall} = \frac{\mathrm{TP}}{\mathrm{TP} + \mathrm{FN}}
\end{equation}

\begin{equation}
\text{F1 Score} = 2 \times \frac{\text{Precision} \times \text{Recall}}{\text{Precision} + \text{Recall}}
\end{equation}

In the context of random and intelligent sampling, we expected that the prediction accuracy of the proposed intelligent scheme should be comparable to or better than that of random sampling. With precision, we aimed to evaluate whether the proposed intelligent sampling method could reduce false positive errors (\textit{i.e.,} the proportion of positive predictions that were actually correct). Similarly, we wanted to quantify the model's ability to accurately specify actual positive cases through recall. An increased recall in the case of intelligent sampling could demonstrate that the model detected a higher number of true cases. Finally, the F1 Score enabled us to determine whether intelligent sampling could offer an improvement over random sampling in assessing false positives and false negatives. From these performance metrics, we expected to ensure the reliability of our approach through the use of better-quality training data.

%-------------------------------
\subsection{Results}
\label{ssec:results}

As mentioned above, we used two models to determine the efficacy of our approach. We created a custom DL model from scratch for training on the coresets. Moreover, we used a pretrained model as a backbone and fine-tuned it for training on the extracted coresets. In particular, we developed the pretrained ResNet-101-v2 model as a backbone. Listing~\ref{lst:nn-model} and Listing~\ref{lst:tl-model} present the details of both these networks. The custom network is a 14-layer CNN architecture, where the final layer is a dense layer with eight outputs specifying the eight PBC classes. Furthermore, Listing~\ref{lst:tl-model} represents the pretrained \textit{ResNet101-v2} used as a base model. We eliminated the top layer of ResNet and kept the weights of the remaining layers. Subsequently, we added five more layers to fine-tune this network on our dataset.

Listing~\ref{lst:nn-model} shows the 14-layered architecture of the neural network that we used in our experiments. The final layer was a dense layer with eight outputs, representing the eight classes that needed to be classified.

Listing~\ref{lst:tl-model} lists the second model that we used for our experiments. This model utilizes pre-trained ResNet101-v2 as a base model pre-trained on the ImageNet dataset. We removed the top layer of ResNet and kept the weights of the remaining layers frozen. Next, we added 5 more layers to fine-tune this network to our dataset. 

\begin{lstlisting}[label=lst:nn-model,caption=Custom Neural Network, basicstyle=\small]
Model: "sequential_9"
______________________________________________________________
 Layer (type)                Output Shape              Param #   
==============================================================
 conv2d_27 (Conv2D)          (None, 254, 254, 32)      896       
 max_pooling2d_27 (MaxPooli  (None, 127, 127, 32)      0         
 ng2D)                                                           
 dropout_36 (Dropout)        (None, 127, 127, 32)      0         
 conv2d_28 (Conv2D)          (None, 125, 125, 16)      4624      
 max_pooling2d_28 (MaxPooli  (None, 62, 62, 16)        0         
 ng2D)                                                           
 dropout_37 (Dropout)        (None, 62, 62, 16)        0         
 conv2d_29 (Conv2D)          (None, 60, 60, 8)         1160      
 max_pooling2d_29 (MaxPooli  (None, 30, 30, 8)         0         
 ng2D)                                                           
 dropout_38 (Dropout)        (None, 30, 30, 8)         0         
 flatten_9 (Flatten)         (None, 7200)              0         
 dense_18 (Dense)            (None, 32)                230432    
 dropout_39 (Dropout)        (None, 32)                0         
 batch_normalization_9 (Bat  (None, 32)                128       
 chNormalization)                                                
 dense_19 (Dense)            (None, 8)                 264       
==============================================================
Total params: 237504 (927.75 KB)
Trainable params: 237440 (927.50 KB)
Non-trainable params: 64 (256.00 Byte)
______________________________________________________________
\end{lstlisting}

\begin{lstlisting}[label=lst:tl-model,caption=Neural Network with ResNet101-v2 Base Model, basicstyle=\small]
Model: "sequential_10"
_____________________________________________________________
 Layer (type)                Output Shape              Param #   
=============================================================
 resnet101v2 (Functional)    (None, 8, 8, 2048)        42626560  
 average_pooling2d_9 (Avera  (None, 1, 1, 2048)        0         
 gePooling2D)                                                    
 flatten_9 (Flatten)         (None, 2048)              0         
 dropout_9 (Dropout)         (None, 2048)              0         
 batch_normalization_9 (Bat  (None, 2048)              8192      
 chNormalization)                                                
 dense_9 (Dense)             (None, 8)                 16392     
=============================================================
Total params: 42651144 (162.70 MB)
Trainable params: 20488 (80.03 KB)
Non-trainable params: 42630656 (162.62 MB)
_____________________________________________________________
\end{lstlisting}

%-------------------------------
\subsubsection{Clustering Results}
\begin{figure}[htbp]
    \centering
    \begin{minipage}{0.45\textwidth}
        \centering
        \includegraphics[width=\textwidth]{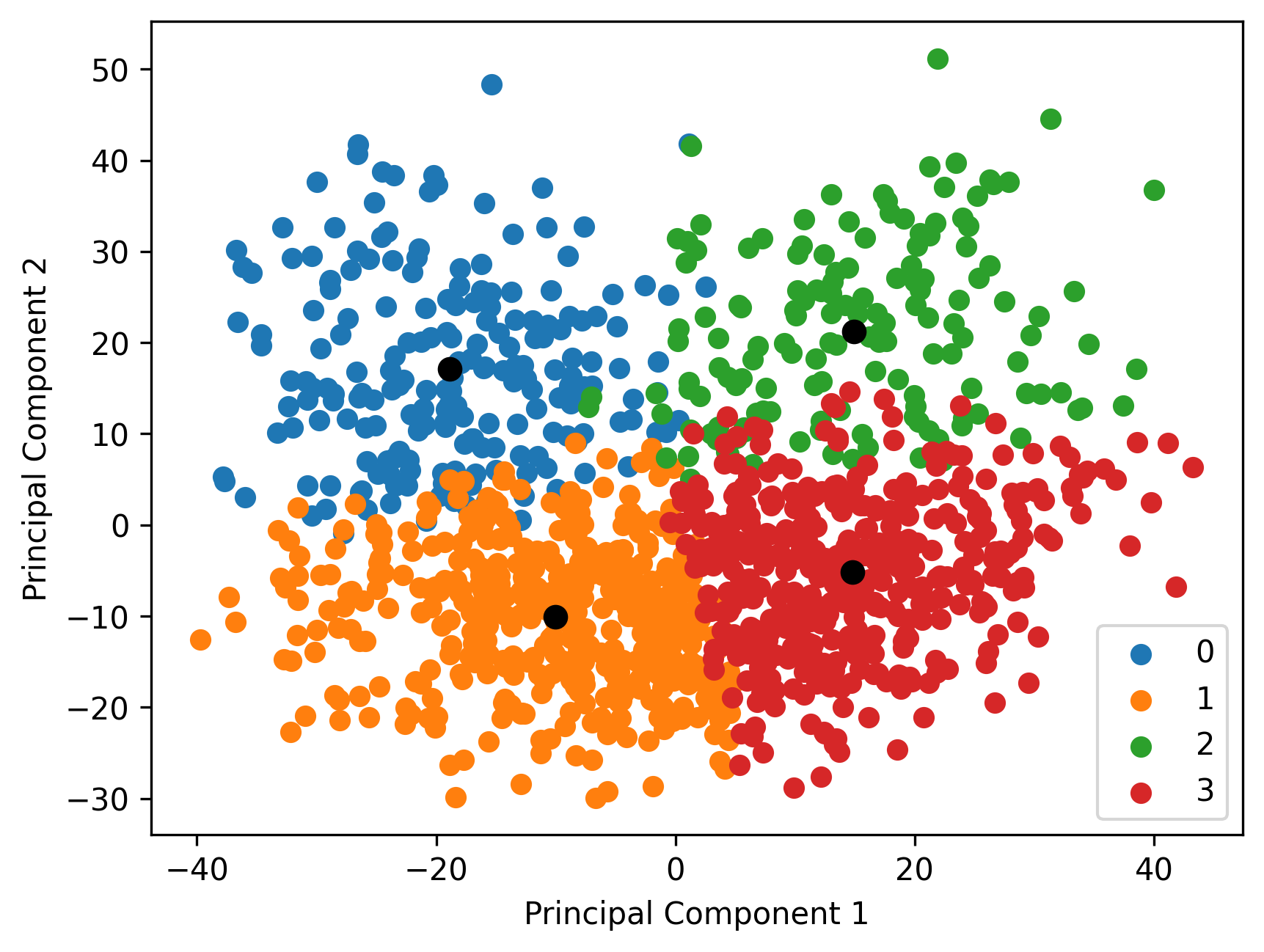}
        \caption{Four interclass clusters were obtained using the $K$-Medoids clustering method within the Lymphocyte class. The black points specify the Medoids or the central data points within a cluster.}
        \label{fig:lym-clusters}
    \end{minipage}\hfill
    \begin{minipage}{0.45\textwidth}
        \centering
        \includegraphics[width=\textwidth]{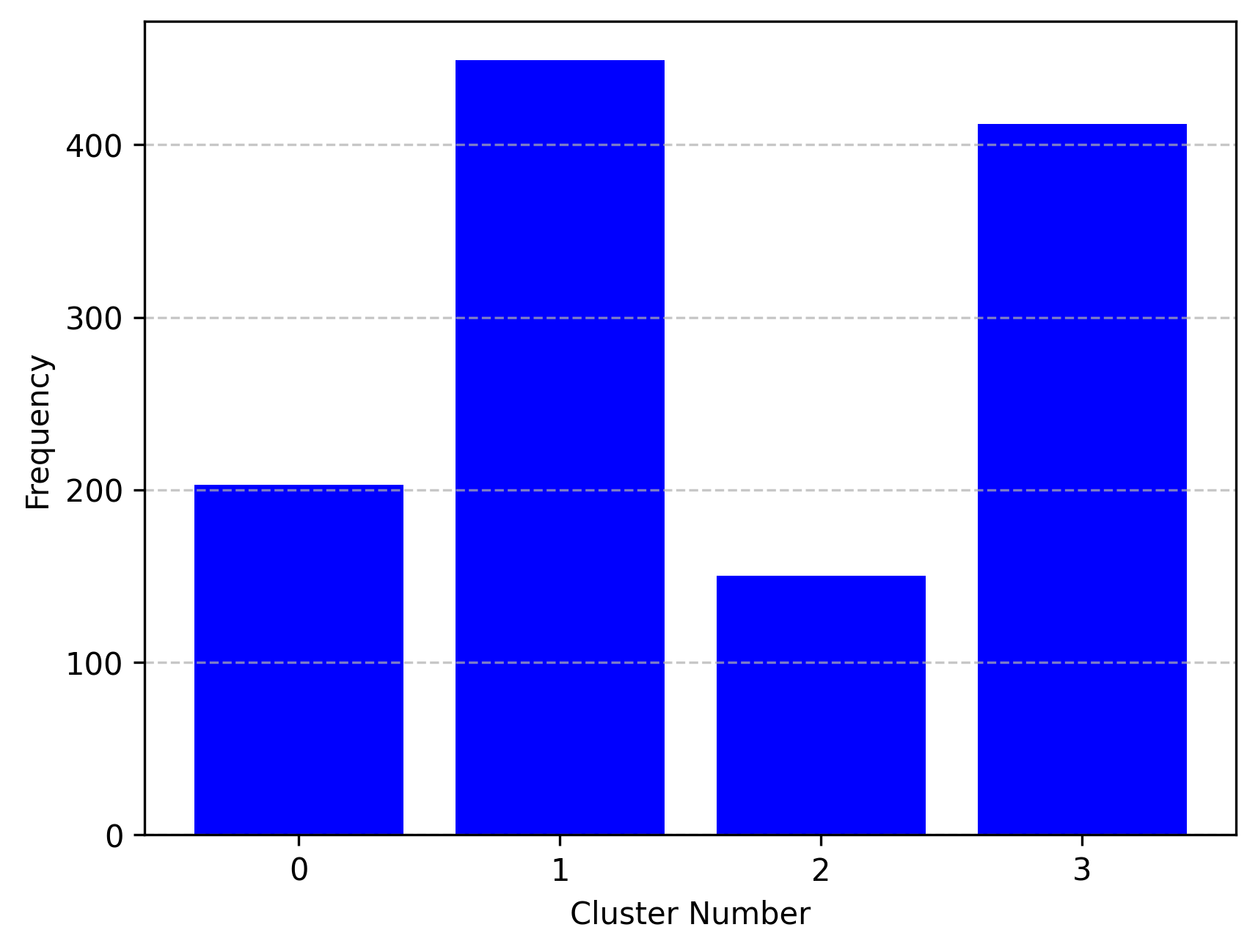}
        \caption{Sample frequency distribution across the four $K$-Medoids clusters of the Lymphocyte class. The diversity of these samples signifies the variability in data points within each cluster, indicating that uniform random sampling may not accurately represent the full diversity.}
        \label{fig:lym-cluster-freq}
    \end{minipage}
\end{figure}

As discussed above, we employed $K$-Medoids clustering to perform \textit{intraclass} splitting of the dataset. Moreover, we utilize the \textit{Silhouette} score to determine the optimal number of clusters. Figure~\ref{fig:lym-clusters} depicts the four clusters identified for the Lymphocyte class. The black markers signify the cluster centroid, which is the Medoids selected by the clustering algorithm.

Figure~\ref{fig:lym-cluster-freq} illustrates the sample distribution based on clusters, revealing that they are not distributed uniformly across clusters. This non-uniform distribution is significant as it signifies a critical limitation of the Random Sampling (RS) technique. More specifically, the RS method results in an inappropriate representation of certain data points within a class. This imbalance in the data representation in a coreset can impact model training, compromising the performance and generalizability of the coreset. In real-world settings, training a DL model on such a coreset can lead to over- or under-fitting of clinically significant cases.

%-------------------------------
\subsubsection{Training Time Results}
\begin{figure}[htbp]
    \centering
    \begin{minipage}{0.48\textwidth}
        \centering
        \includegraphics[width=\textwidth]{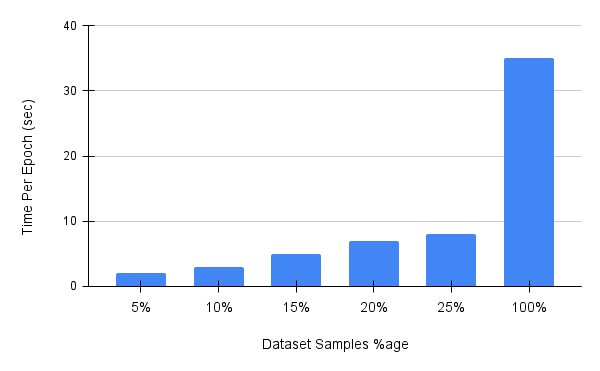}
        \caption{Training time (in seconds) for the custom DL model as a function of training coreset size (expressed as percentages of the total dataset). Training time increases as the coreset size increases.}
        \label{fig:train-time-cn}
    \end{minipage}\hfill
    \begin{minipage}{0.48\textwidth}
        \centering
        \includegraphics[width=\textwidth]{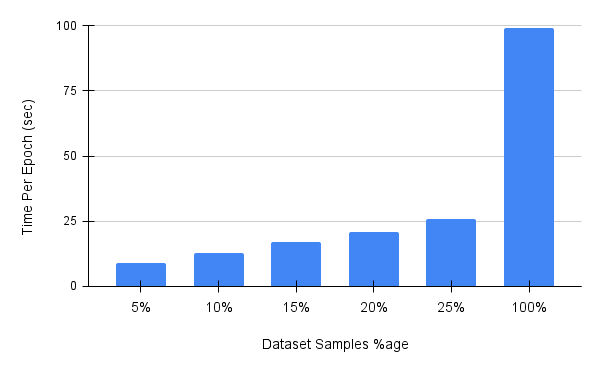}
        \caption{Training time (in seconds) per epoch for the neural network trained using ResNet101-v2 backbone across various training coreset sizes. As expected, these results demonstrate a trend of increased computational costs when the coreset size increases.}
        \label{fig:train-time-tl}
    \end{minipage}
\end{figure}

We determined the training time for both the customized DL model built from scratch and the ResNet model. These results are presented in Figure~\ref{fig:train-time-cn} and Figure~\ref{fig:train-time-tl}. On the x-axis, we plot the coreset size as a percentage of the total dataset, while presenting training time (in seconds) on the y-axis. As expected, the training time of both models increases as the coreset size increases. These results align perfectly with the theoretical expectation that a large dataset requires more computational resources and convergence time, regardless of the physical architecture of the model.

%-------------------------------
\subsection{Performance Results}
This section summarizes experimental results evaluating the performance of two models in comparison to the proposed schemes. The dataset was segregated for these experiments using an 80/20 split between training and validation datasets. As discussed above, we employed two types of network architectures: a customized neural network developed from scratch, trained on our dataset, and a pretrained ResNet101-v2-based model trained on the ImageNet dataset. The latter was tailored to our problem through transfer learning methods. ~\ref{ssec:results} elaborates on the architectures of both networks.

\begin{figure}
 \begin{subfigure}{0.49\textwidth}
     \includegraphics[width=\textwidth]{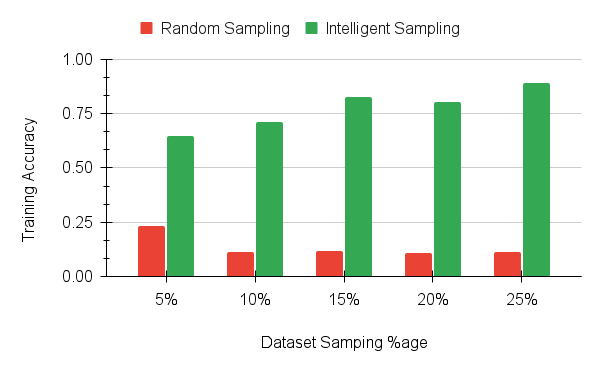}
     \caption{Training accuracy of the customized neural network built from scratch for various coreset sizes for both RS and IS methods, where the former exhibits underfitting and the latter accomplishes enhanced accuracy, even for small coreset sizes.}
     \label{fig:a}
 \end{subfigure}
 \hfill
 \begin{subfigure}{0.49\textwidth}
     \includegraphics[width=\textwidth]{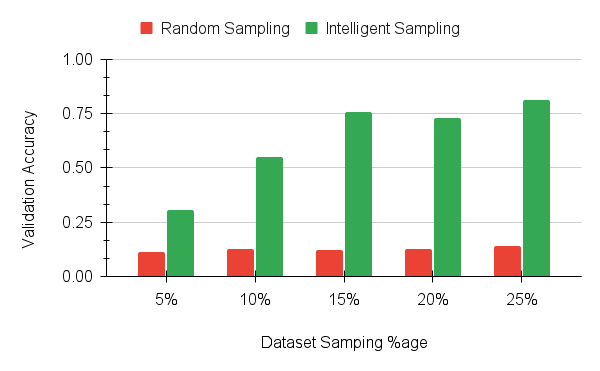}
     \caption{Validation accuracy results of the customized neural network developed from scratch for various coreset sizes for both RS and IS methods, where IS constantly outperforms RS, even for smaller coreset sizes.}
     \label{fig:b}
 \end{subfigure}
 
 \medskip
 \begin{subfigure}{0.49\textwidth}
     \includegraphics[width=\textwidth]{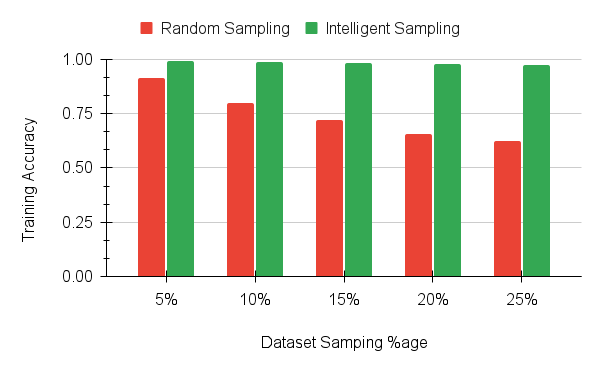}
     \caption{Training accuracy results of the ResNet-based neural network trained with various coreset sizes, where the IS-based method outperforms the RS method. However, RS provides high training accuracy, exhibiting potential overfitting.}
     \label{fig:c}
 \end{subfigure}
 \hfill
 \begin{subfigure}{0.49\textwidth}
     \includegraphics[width=\textwidth]{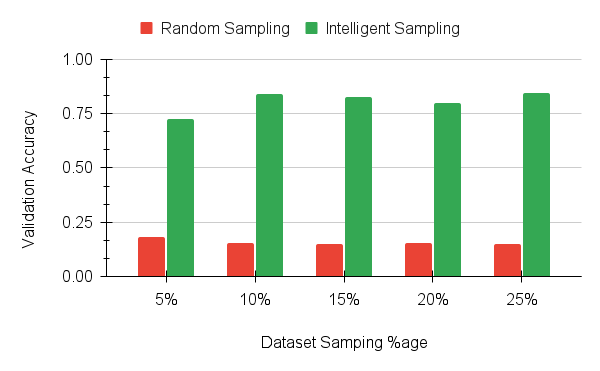}
     \caption{Validation accuracy results of the ResNet-based neural network trained with different coreset sizes, where the IS-based method shows better generalization and outperforms the RS-based method for all coreset sizes.}
     \label{fig:d}
 \end{subfigure}

 \caption{Training and Validation Accuracy Results for Random and Intelligent Sampling}
 \label{fig:acc-results}

\end{figure}

Figure~\ref{fig:acc-results} presents the performance results to demonstrate the accuracy of both the training and validation datasets for the two networks. We used diverse coreset sizes to produce results for two types of sampling, which have been discussed as follows:

\begin{itemize}
    \item Random Sampling (RS): Randomly selected samples from the dataset.
    \item Intelligent Sampling (IS): Samples selected through the proposed intelligent sampling technique.
\end{itemize}

The results presented in Figure~\ref{fig:a} demonstrate that the proposed IS-based scheme outperformed the RS-based method across both networks. In the case of a customized network, the RS-based strategy leads to underfitting, as specified by the poor training accuracy results (<$25\%$ even for a $5\%$ sample) given in Figure~\ref{fig:a}. In contrast, the IS-based approach offers $65\%$ training accuracy even for a coreset representing $5\%$ of the dataset. In addition, as the coreset size gradually increases, the RS-based scheme underperforms, whereas the IS-based method improves the accuracy further.

In the case of ResNet-based architecture, the model trained on the RS-based coreset overfits, as illustrated by decent training accuracy results in Figure~\ref{fig:c}. However, it has poor validation accuracy, as shown in Figure~\ref{fig:d}. Even with better training accuracy, the results have a decreasing trend as the coreset size increases. On the other hand, the model trained on the coreset selected by the proposed IS-based scheme demonstrated consistently higher accuracy for both training and validation datasets, as depicted in Figure~\ref{fig:c} and Figure~\ref{fig:d}.

\begin{table}[!htp]
\centering
\caption{A summary of various performance metrics for two models across three dataset configurations: $100\%$ dataset, $25\%$ RS-based coreset, $25\%$ IS-based coreset. Columns 4 and 7 show that both models trained on IS-based coresets perform comparably to these models trained on the $100\%$ dataset.  }
\label{tab:perf-results}
\scriptsize
\begin{tabular}{lrrr|rrrr}\toprule
&\multicolumn{3}{c}{\textbf{Custom Network Results}} &\multicolumn{3}{|c}{\textbf{Transfer Learning Results}} \\\cmidrule{2-7}
\textbf{Metric} &\textbf{100\%} &\textbf{RS (25\%)} &\textbf{IS (25\%)} &\textbf{100\%} &\textbf{RS (25\%)} &\textbf{IS (25\%)} \\\midrule
\textbf{Accuracy}   & 0.865007  & 0.125     & 0.813049  & 0.865007  & 0.125     & 0.813049 \\
\textbf{Precision}  & 0.857411  & 0.015625  & 0.817497  & 0.857411  & 0.015625  & 0.817497 \\
\textbf{Recall}     & 0.860387  & 0.125     & 0.812285  & 0.860387  & 0.125     & 0.812285 \\
\textbf{F1 Score}   & 0.855442  & 0.027778  & 0.797999  & 0.855442  & 0.027778  & 0.797999 \\
\bottomrule
\end{tabular}
\end{table}

\begin{figure}[htbp]
    \centering
    \begin{minipage}{0.48\textwidth}
        \centering
        \includegraphics[width=\textwidth]{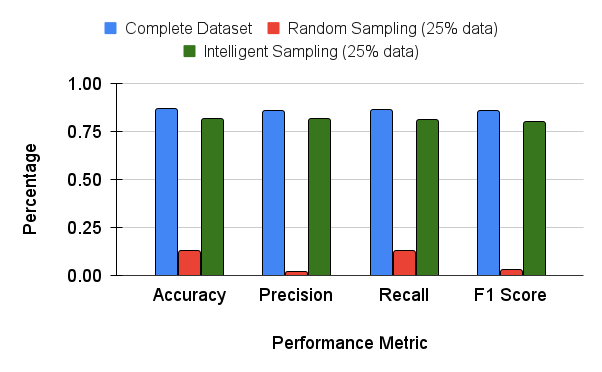}
        \caption{Evaluation of Precision, Recall, and F1 Score for customized network across $100\%$ dataset, $25\%$ RS, and $25\%$ IS training configurations.}
        \label{fig:perf-results-cn}
    \end{minipage}\hfill
    \begin{minipage}{0.48\textwidth}
        \centering
        \includegraphics[width=\textwidth]{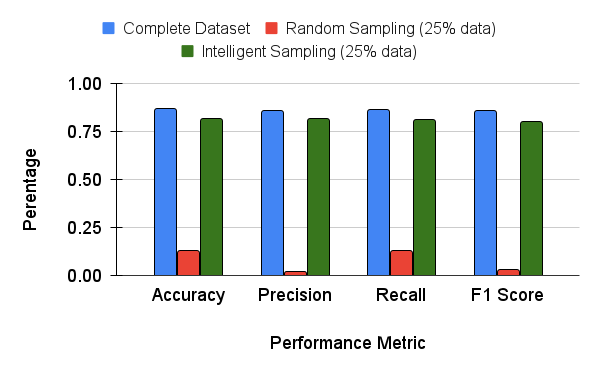}
        \caption{Evaluation of Precision, Recall, and F1 Score for the neural network trained using ResNet101-v2 Backbone across $100\%$ dataset, $25\%$ RS, and $25\%$ IS training configurations.}
        \label{fig:perf-results-tl}
    \end{minipage}
\end{figure}

Besides accuracy, we also produced vital results on other metrics, such as Precision, Recall, and F1 Score,  to evaluate our models more thoroughly. Table~\ref{tab:perf-results} presents these results, which are also depicted in Figure~\ref{fig:perf-results-cn} and Figure~\ref{fig:perf-results-tl}. We reported performance metrics for models trained on the $100\%$ dataset to create a baseline comparison. For the sake of brevity, Figure~\ref{fig:perf-results-cn} and Figure~\ref{fig:perf-results-tl} show only results for a $25\%$ coreset size for both RS- and IS-based methods to streamline presentation.

The results reveal a significant disparity between the RS-based and the IS-based methods. We can observe that the RS-based coreset yielded poor results across all performance metrics, showing its limited practicality in real-world scenarios. On the other hand, the IS-based coreset yielded excellent results for all metrics, including Precision, Recall, and F1-score. In Table~\ref{tab:perf-results} (columns 4 and 7), we can observe that this trend is consistent for both customized neural network and ResNet-based models in the case of the IS method.

Remarkably, the most significant result is that the model trained on the IS-based coreset (only $25\%$ of the entire dataset) achieves nearly equivalent performance to the model trained on the full dataset. We can observe these results while comparing columns 2 and 4 in Table~\ref{tab:perf-results} for the customized network and columns 5 and 7 for the ResNet-based model. Furthermore, Figure~\ref{fig:perf-results-cn} and Figure~\ref{fig:perf-results-tl} also depict these outcomes, demonstrating the effectiveness of the proposed IS-based approach in specifying representative data points within a large dataset. We believe that if the proposed method can produce comparable performance for only $25\%$ of the whole dataset, it can significantly reduce computational costs without compromising the quality of outcomes. Consequently, it can produce a transformative impact in real-world scenarios where computational efficiency has become essential in designing advanced predictive models.

%%%%%%%%%%%%%%%%%%%%%%%%%%%%%%%%
\section{Conclusion and Future Work}

This study examines the effectiveness of Deep Learning (DL) models in classifying blood cells using the PBC image dataset. In particular, we introduced an innovative intelligent sampling methodology that captures intraclass diversity to improve training efficiency and model generalizability. We observed that training a DL model on IS-based coresets provides two significant advantages: first, it minimizes the overfitting risk by discarding redundant data points or those without relevant information; second, it reduces computational costs during model training and hyperparameter optimization.

The experiment results demonstrate that the random sampling method is ineffective in producing an effective training coreset for both customized and ResNet-based models, resulting in suboptimal performance. In contrast, the proposed IS-based technique could identify a representative coreset (only $25\%$ of the whole dataset) that produced comparable performance to the full dataset when both models were trained on two configurations of datasets and evaluated across several metrics. These outcomes demonstrate the efficacy of the IS-based method in producing compressed yet representative training coresets. As a result, such coresets can be highly beneficial in resource-efficient DL model training by ensuring remarkable computational cost reduction while maintaining high classification accuracy and robustness.

In the future, we will perform an ablation study to evaluate the performance of the IS-based technique and determine the impact of various parameters, including silhouette scoring and cluster count, on different metrics. The current study is based on a single biomedical imaging dataset, PBC, which limits the generalizability of the proposed method. Therefore, we will test our method on various other imaging datasets, such as dermatology or histopathology scans. Furthermore, we will expand the idea of IS-based coreset selection for other non-medical classification studies. 

%%%%%%%%%%%%%%%%%%%%%%%%
\section*{Data Availability}
The authors utilized publicly available Peripheral Blood Cell (PBC)~\cite{pbc-dataset} dataset for experiments.

%%%%%%%%%%%%%%%%%%%%%%%%
\bibliographystyle{alpha}
\bibliography{refs.bib}

% ------------------------------------------------------------
% ============================================================
% ------------------------------------------------------------

\end{document}